\documentclass{article}

\usepackage[final, nonatbib]{neurips_2023}

\usepackage[square,sort,comma,numbers]{natbib}

\usepackage[utf8]{inputenc} 
\usepackage[T1]{fontenc}    
\usepackage{hyperref}       
\usepackage{url}            
\usepackage{booktabs}       
\usepackage{nicefrac}       
\usepackage{microtype}      

\usepackage{amsmath,amssymb,amsfonts}
\usepackage{algorithmic}
\usepackage{graphicx}
\usepackage[figurename=Fig.,font=small,skip=2pt]{caption}
\usepackage{textcomp}
\usepackage{enumerate}
\usepackage{xcolor}
\usepackage{tabularray}
\usepackage{multirow}

\title{A Novel Vision Transformer with Residual in Self-attention for Biomedical Image Classification}

\author{%
  Arun K. Sharma \\
  Dept. of Electrical Engineering\\
  Indian Institute of Technology\\
  Kanpur, India \\
  \texttt{email}: arnksh@iitk.ac.in\\
  \And
  Nishchal K. Verma \\
  Dept. of Electrical Engineering \\
  Indian Institute of Technology \\
  Kanpur, India\\
  \texttt{email:} nishchal@iitk.ac.in\\
}

\begin{document}

\maketitle

\begin{abstract}
 Biomedical image classification requires capturing of bio-informatics based on specific feature distribution. In most of such applications, there are mainly challenges due to limited availability of samples for diseased cases and imbalanced nature of dataset. This article presents the novel framework of multi-head self-attention for vision transformer (ViT) which makes capable of capturing the specific image features for classification and analysis. The proposed method uses the concept of residual connection for accumulating the best attention output in each block of multi-head attention. The proposed framework has been evaluated on two small datasets: (i) blood cell classification dataset and (ii) brain tumor detection using brain MRI images. The results show the significant improvement over traditional ViT and other convolution based state-of-the-art classification models.
\end{abstract}

\section{Introduction}
\label{intro}
With the advancement of modern computational intelligence, biomedical analysis has evolved enormously and has become capable of capturing several biomedical informatics from patients \cite{ch6r1}. The biomedical images obtained from various medical imaging modalities, such as radiology, histology, microscopy, etc., can be used to train an intelligent system capable of helping medical specialists diagnose various diseases using medical imaging. However, there are various challenges in training deep learning models for biomedical image data classification and analysis due to limited availability, mostly imbalanced data, and the need for medical specialists for image labeling. Many researchers have put their efforts into meeting such challenges and developing versatile and efficient models for disease diagnosis based on biomedical imaging \cite{ch6r2_0, ch6r2, ch6r3, ch6r4, ch6r6}.

Traditional machine learning methods, such as support vector machines (SVMs) and multi-layer perceptrons (MLPs), have long been utilized for biomedical image classification \cite{ch6r2_0}. However, these methods have various drawbacks. Firstly, their performance falls short of practical standards, and their development has been relatively slow in recent years. Further, the process of feature extraction and selection in these methods is time-consuming and varies depending on the objects being classified \cite{ch6r7}. In contrast, deep neural networks (DNNs), such as various models based on convolutional neural networks (CNNs), have gained widespread adoption in biomedical image data classification and analysis \cite{ch6r5}. For instance, CheXNet, a CNN model with 121 layers was trained on a dataset with $10^5$ samples of frontal-view chest X-rays (ChestX-ray 14). Its performance was superior to the average performance of four radiologists. The main issue in training a DL model for biomedical image data classification was the limited availability of labeled images for diseased cases.  To resolve this issue, a number of studies have reported the application of transfer learning (TL) to train the model faster with limited availability of image samples \cite{ch6r6, ch6r7, ch6r8}. By training the DL models using TL, Daniel. et. al. \cite{ch6r7} achieved 92\% of accuracy in classifying X-ray images after training with small training samples.  These methods further enhance the success of biomedical image classification and analysis using machine learning rather than human experts. 

However, there are still some research gaps that need to be addressed \cite{ch6r9}. For example, the problem of overfitting exits for almost all CNN-based models due to the variability and heterogeneity of the biomedical image dataset. Biomedical images can exhibit significant variability and heterogeneity in terms of patient characteristics, imaging protocols, imaging modalities, and disease manifestations. These make it challenging for models to generalize across different scenarios and accurately capture relevant features.

The objective of this chapter is to investigate and develop an advanced deep learning model that can generalize across different scenarios of medical imaging and accurately capture relevant features for classification and diagnostics. Tiago et. al \cite{vitm2} has discussed the importance and effect of attention mechanism for medical image classification. Vision transformer (ViT) by Alexey et. al. \cite{vit} was the first to use the attention mechanism for image classification inspired by transformer architecture for natural language processing (NLP). The vision transformer has been reported to outperform the CNN based models if trained with large number of images \cite{vit}. In recent years, there have been various improvements in vision transformer such as data-efficient image transformers \& distillation through attention \cite{vitm01}, hybridization with CNN model \cite{vitm0}, and swin transformer \cite{swinvit}. Several studies have been reported on the application of vision transformer in biomedical image classification and analysis in different scenario \cite{vitm3, vitm4, vitm5, vitm6, vitm7}. 

H. Wang et. al. \cite{vitm3} introduces multi-stage convolutional layer based ViT-Plus model for the classification of genitourinary syndrome of menopause using OCT images. They utilised multi-stage Convo2d layer to convert the original images into patches rather than using single Convo2d layer.  Mashood et. al. \cite{vitm4} presents the application of vision transformer and language model for the generation of radiology report images. Sabry et. al. \cite{vitm5} introduced a hybrid mechanism using autoencoder, infoGAN and vision transformer for image retrieval in unsupervised manner. Ren et. al. \cite{vitm6} presents the application of vision transformer for COVID-19 case detection using X-ray and CT image classification. They reported to achieve the accuracy of 97.65\% and 99.12\% for X-ray and CT image dataset respectively. Manzari et. al introduced a robust medical vision transformer by considering hybridization of convolutional, transformer and its augmentation block for medical image classification. Based on the various study reported on different variants of ViT and their application for image classification, here we present the residual self-attention in ViT for biomedical image classification. To the best of our knowledge, there is no study reported on multi-head self-attention with residual connection using best probabilistic attention output. The highlights of the contributions of this chapter has been listed below:
\begin{enumerate}
    \item This chapter introduces a new multi-head attention mechanism using residual connection from the best of probabilistic attention.
    \item The Manhattan norms, also known as the L1 norms of the attention score (probabilistic output of Key and Query) from all the attention heads are computed. 
    \item The best of the L1 norms of the attention scores is used as residual connection to the final projected attention output.
    \item The proposed method is applied for the classification of biomedical images from (i) blood cell images (class names: Eosinophil, Lymphocyte, Monocyte, and Neutrophil) and (ii) brain tumer detaction using brain MRI images
\end{enumerate}

\section{Theoretical Background}\label{sec:theo_vit}
\begin{figure}[!ht]
\centering
\includegraphics[width=10cm]{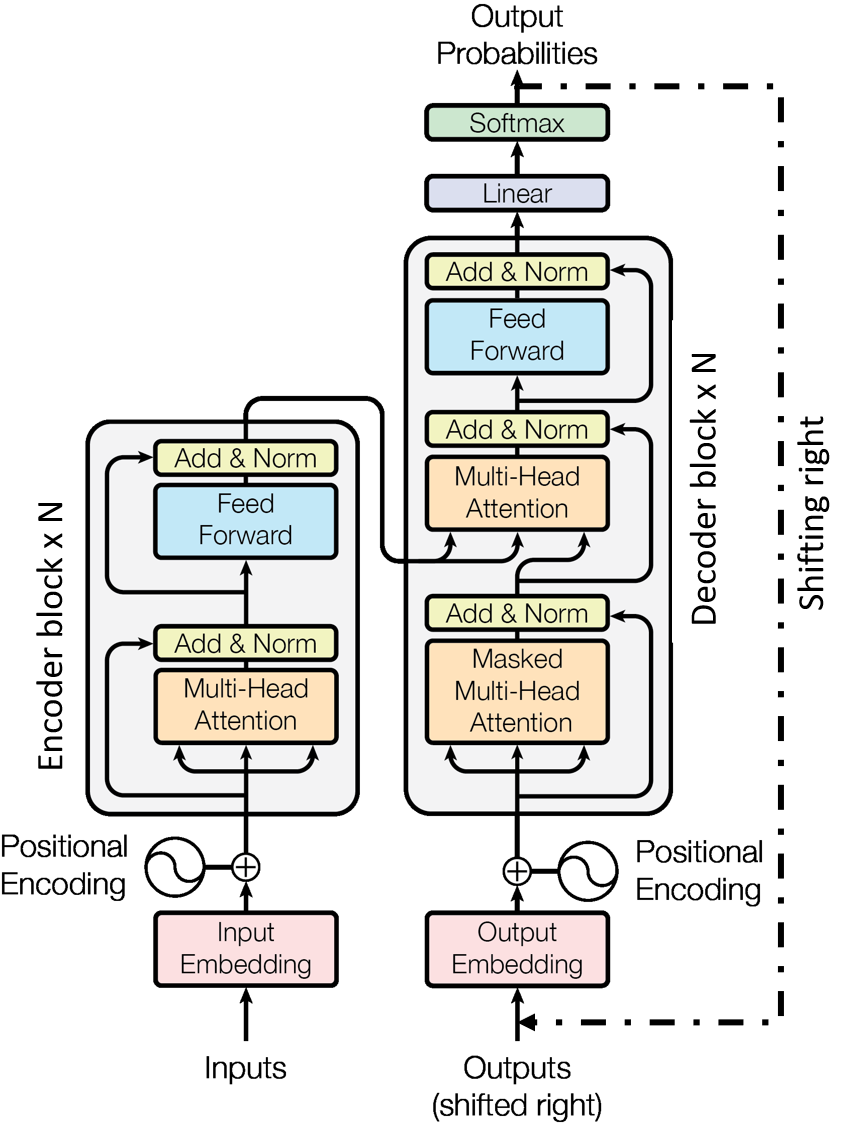}
\caption{Architecture of Transformer in NLP \cite{trans01}}
\label{ch_vit:trans_nlp}
\end{figure}

\subsection{Transformer in NLP} \label{sec:trans_nlp}
The transformer introduced in the seminal paper "Attention is All You Need" by Vaswani et. al. \cite{trans01} is a powerful architecture that revolutionized natural language processing (NLP) tasks. The transformer architecture has since become the foundation for various state-of-the-art NLP models as well as computer vision models. It leverages the self-attention mechanism to efficiently capture the inter-dependencies in long sequential inputs. This mechanism enables the model to consider the context of each word based on the entire input sequence, resulting in better representation learning.

The schematic of transformer architecture has been shown in Fig. \ref{ch_vit:trans_nlp}. The main components of the transformer are encoder and decoder. The encoder processes the input sequence, while the decoder generates an output sequence. Each encoder and decoder layer in a transformer contains multiple self-attention heads, allowing the model to attend to different parts of the input sequence simultaneously. The traditional recurrent models, like RNNs, suffer from vanishing or exploding gradients when dealing with long sequences. On the other hand, transformers overcome this limitation by employing attention mechanisms, allowing them to capture dependencies between distant words.

\subsection{Vision Transformer} \label{sec:vit}
With the success of transformer model in NLP, the attention based architecture has been extended to a ground-breaking architecture called Vision Transformer (ViT).  Introduced in the paper "An Image is Worth 16x16 Words: Transformers for Image Recognition at Scale" by Dosovitskiy et al. in 2020 \cite{vit}, it has been reported to achieve remarkable results in image classification tasks. Unlike traditional CNN models, which have been the dominant approach in computer vision, ViT relies solely on the self-attention mechanism of transformers to capture relationships between image patches.
\begin{figure}[!ht]
\centering
\includegraphics[width=15cm]{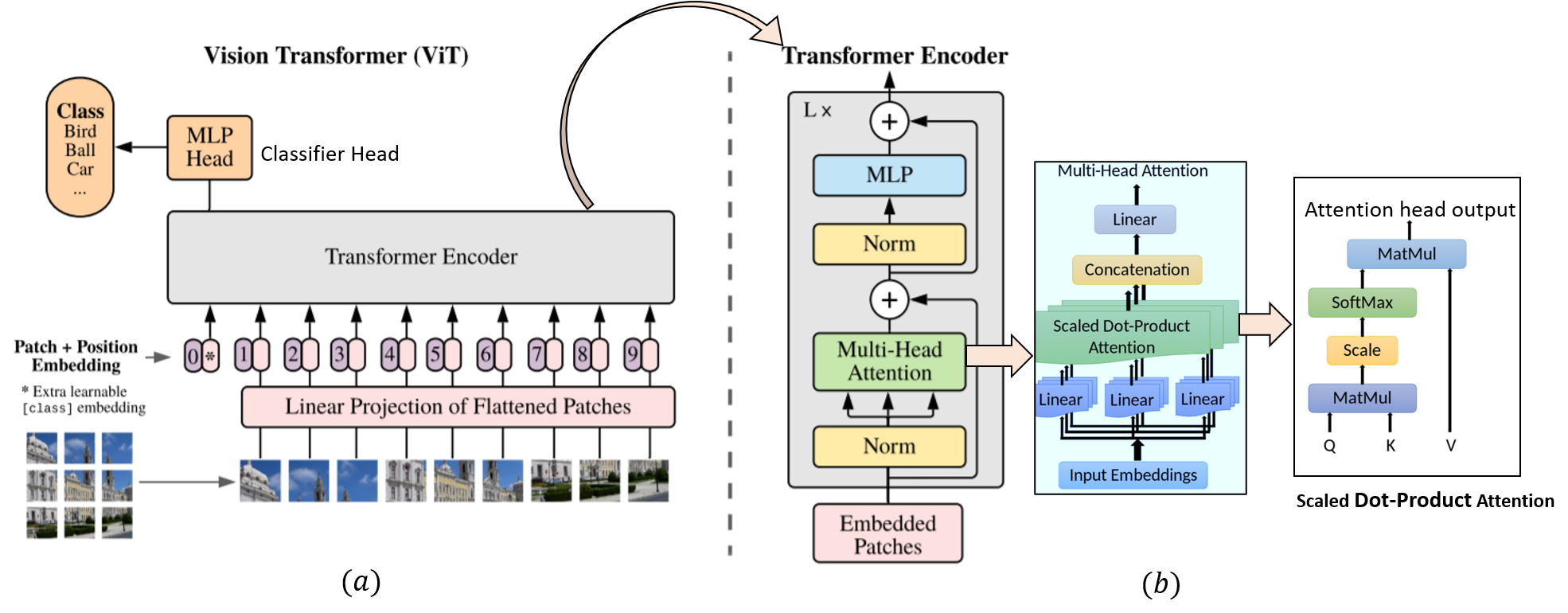}
\caption{Architecture of Vision Transformer for computer vision \cite{vit}}
\label{ch_vit:vit}
\end{figure}

The schematic architecture of vision transformer and multi-head attention mechanism has been depicted in Fig. \ref{ch_vit:vit}. Contrary to transformer architecture shown in Fig. \ref{ch_vit:trans_nlp}, the vision transformer consists of only encoder block which accepts input image as a sequence of fixed-size patches. These patches are then linearly embedded into tokens with positional encoding similar to word embedding in NLP to capture the visual information of the input image. The image patches are then processed by the encoder block, which consists of multiple heads of self-attention and feed-forward neural networks (as shown in Fig. \ref{ch_vit:vit}(b)). The key components of ViT are explained briefly as follows
\begin{enumerate}[i)]
    \item \textbf{Position embedding} It captures the spatial information within the input image. Since the transformer architecture does not inherently possess knowledge of spatial relationships, positional embeddings are provided to the understanding of patch positions within the image. The positional embeddings are usually fixed, representing the spatial coordinates of each patch. They are added to the token embeddings, allowing the model to differentiate between patches based on their relative positions.
    \item \textbf{Multi-Head Attention} It consists of multiple heads of self-attention block. Each attention block receives three inputs: Query (Q), Key (K), and Value (V) generated by three different linear layers with learnable parameters. Then the three terms Q, K, and V are processed by scaled dot attention block as described in equation (\ref{eq:scaled_dot}.
        \begin{equation}\label{eq:scaled_dot}
            Output = SoftMax\left(\frac{Q.K^{T}}{\sqrt{d}}\right)V
        \end{equation}
        where, $d$ represents the size of attention head (number of hidden layers in the linear layer in attention block). The attention output from all the attention heads are concatenated and projected by a linear layer to its original size.
    \item \textbf{Feed-forward Neural Networks} Feed-forward neural networks are provided with residual connection to process the token embeddings and extract higher-level features from the image patches based on the attention output. Usually two linear layer with a non-linear activation function in between them are used to construct the MLP block. The first linear layer applies a linear transformation to the input features, mapping them to a higher-dimensional space. This linear transformation is parameterized by a weight matrix and a bias term. Next, a non-linear activation function is applied element-wise to the transformed features. This introduces non-linearity into the model and allows it to capture more complex relationships between the input features. After the activation function, the output is passed through the second linear layer, which maps the features back to the original embedding dimension.
    \item \textbf{Classifier Head:} Classifier head is a multi-layer neural network with SoftMax layer at the end to process the output features from the transformer encoder and produce the class decision.
\end{enumerate}

\section{Proposed Vision Transformer}\label{sec:proposed} 
\begin{figure}[!ht]
\centering
\includegraphics[width=15cm]{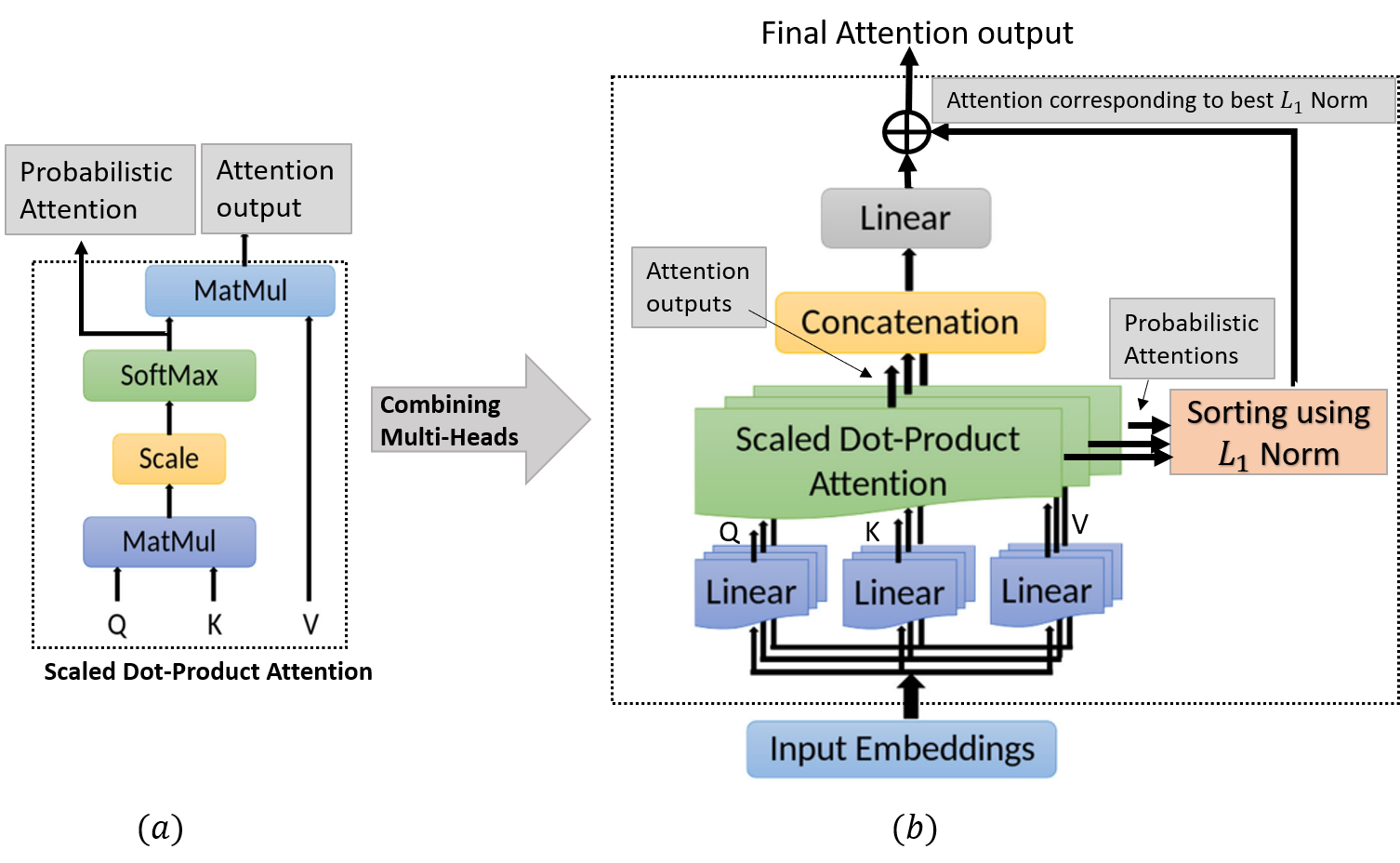}
\caption{The proposed attention mechanism in Vision Transformer}
\label{ch_vit:proposed}
\end{figure}

In this section, our proposed vision transformer is presented. Since, the attention is the major component of ViT that captures the relationship between the sequential input patches, the objective of this section is to present the improved mechanism of multi-head self-attention. The proposed novel multi-head self-attention is illustrated in Fig. \ref{ch_vit:proposed} and is explained in following steps.

\begin{enumerate}[1)]
    \item The scaled Dot-Product attention block (Fig. \ref{ch_vit:proposed}(a)) is made to produce two outputs: probabilistic attention which is the softmax of scaled value of matrix multiplication of $Q$ \& $K^T$ and attention output which is the matrix multiplication of probabilistic attention and the value ($V$). Both the outputs from all such attention heads are collected.
    \item The attention outputs are concatenated and passed through a linear projection layer as in the original version of ViT. 
    \item $L_1$ norms of the probabilistic outputs from all attention heads are computed and sorted. The probabilistic attention with highest norm value is treated as attention head having more attention probability. The attention output ($MatMul$ of probabilistic attention term and the value ($V$) term) corresponding to the best attention probability is collected. This attention output is added as residual term to final projected attention output by the linear projection layer as shown in Fig. \ref{ch_vit:proposed}(b).
\end{enumerate}

\section{Experimental Results and Discussion}\label{sec:vit_expD} The efficacy of the proposed framework of novel attention mechanism is demonstrated in this section for (i) blood cells classification dataset\cite{bccd} and (ii) brain tumor detection using MRI images \cite{mri_data}.

\subsection{Dataset Description}
\subsubsection{Blood cell image dataset } 
\begin{figure}[!ht]
\centering
\includegraphics[width=12 cm]{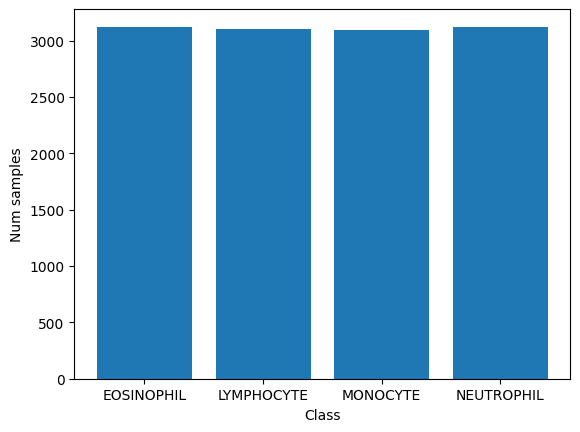}
\caption{Class distribution of the blood cell image dataset}
\label{fig:bccd_distr}
\end{figure}

Blood cell image dataset, also known as Blood Cell Count and Detection (BCCD) dataset has been taken from publicly available repository at Kaggle \cite{bccd}.  It is the collection of 12,500 images of blood cells of four different type: Eosinophil, Lymphocyte, Monocyte, and Neutrophil. Each category contains 3000 samples placed in different folders namely: Eosinophil, Lymphocyte, Monocyte, and Neutrophil. This dataset is very useful for classification of blood cell sub-types which ultimately helps for characterising the blood cells for the diagnosis of blood-based diseases. Class distribution of the dataset has been illustrated in Fig. \ref{fig:bccd_distr}. The available samples are further split into training and testing set in ratio 80:20. Therefore, the train and test sets contain 2400 and 600 samples per class respectively. 

\subsubsection{Brain MRI image dataset} 
\begin{figure}[!ht]
\centering
\includegraphics[width=10 cm]{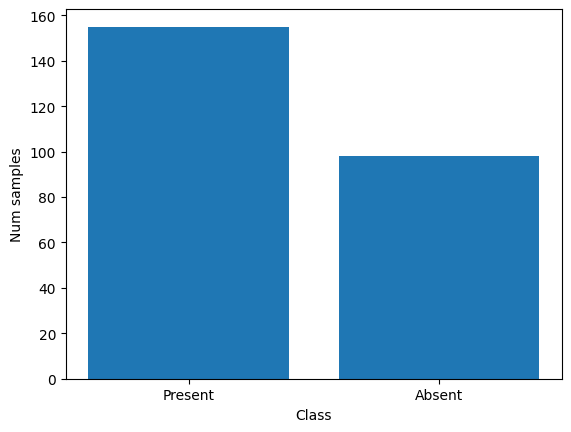}
\caption{Class distribution of Brain MRI Images for Brain Tumor Detection}
\label{fig:mri_distr}
\end{figure}

Brain MRI Images for Brain Tumor Detection has been taken from publicly available repository Kaggle \cite{mri_data}. This is small dataset that contains Magnetic Resonance Imaging (MRI) images of brain which represents patients having brain tumor and having no tumor. Class-wise distribution of samples for both classes is depicted in Fig. \ref{fig:mri_distr}. It can be observed that the dataset is highly imbalanced with 155 of samples for brain tumor cases and 98 samples of healthy (no tumor) cases. The available samples are further split into training and testing set in ratio 80:20. Therefore, the train sets contains 128 samples for tumor cases and 74 samples of healthy cases. Similarly, the test set contains 27 samples in tumor cases and 24 samples of healthy cases 

\subsection{Implementation Details} The proposed ViT with novel multi-head self-attention mechanism as shown in Fig. \ref{ch_vit:proposed} is implemented using PyTorch library of python on Google Co-laboratory platform. Since, training of ViT from scratch requires large number of samples, but both the datasets considered in this study are small dataset, pre-trained ViT model ``\textbf{B\_16\_imagenet1k}'' has been used for implementation.  The ViT model ``\textbf{B\_16\_imagenet1k}'' has been pre-trained on ImageNet-21k dataset and further fine tuned on ImageNet-1k dataset. The implementation of the proposed attention mechanims with the existing pre-trained model has described in following steps:
\begin{enumerate}[i)]
    \item The multi-head self-attention with residual connection as depicted in Fig. \ref{ch_vit:proposed} is implemented as Python class definition using nn-module of PyTorch.
    \item ViT model with pre-trained weights is loaded
    \item The multi-head attention block is replaced with proposed attention implementation. 
    \item The weights of the proposed block is initialised with random initialization.
\end{enumerate}

\subsection{Evaluation Metrics} As the dataset for the brain tumor detection using brain MRI images is highly imbalanced, performance of the model is evaluated using classification accuracy ($CA$), precision, recall, F-1 score, and confusion matrix. 

\subsection{Comparison with State-of-the-art Methods} The proposed framework of ViT with new multi-head self-attention unit is compared with the state-of-the-art methods for image classification. The state-of-the-art methods selected are listed below
\begin{enumerate}[i)]
    \item Convolutional Neural Network (CNN) with custom architecture,
    \item AlexNet pre-trained model and fine-tuned on the current dataset \cite{alexnet},
    \item ResNet18 pre-trained and fine-tuned on the current dataset \cite{resnet50},
    \item Pre-trained ViT ``\textbf{B\_16\_imagenet1k}'' and fine tuned on the current dataset \cite{vit}.
\end{enumerate}

\subsection{Results}
\begin{figure}[!ht]
   \centering
   \begin{minipage}{0.49\textwidth}
    \includegraphics[width=0.97\textwidth]{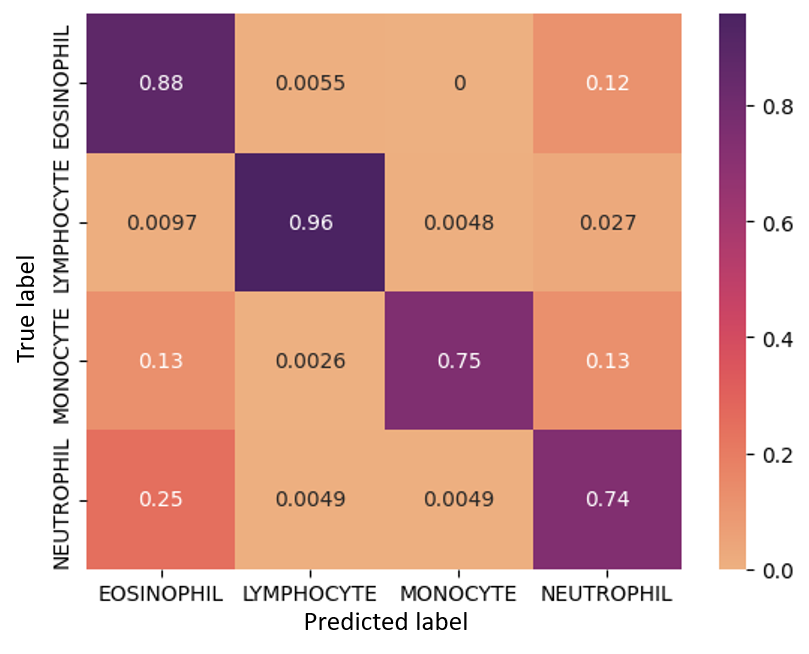}
    \caption{ViT pre-trained for BCCD data.}
    \label{fig:conf_cnn}
\end{minipage}
\begin{minipage}{0.49\textwidth}
   \centering
    \includegraphics[width=0.97\textwidth]{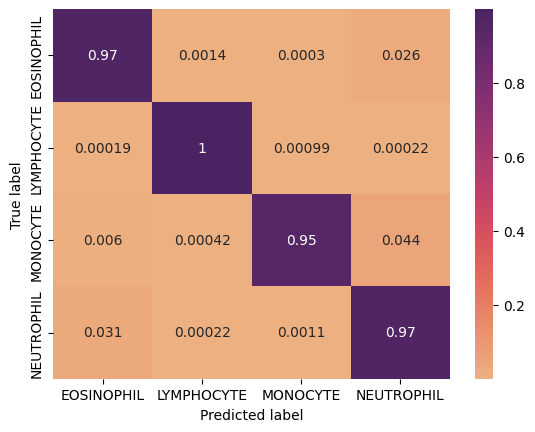}
    \caption{ViT Proposed for BCCD data.}
    \label{fig:conf_alexnet}
    \end{minipage}
\end{figure}

\begin{figure}[!ht]
   \centering
   \begin{minipage}{0.49\textwidth}
    \includegraphics[width=0.97\textwidth]{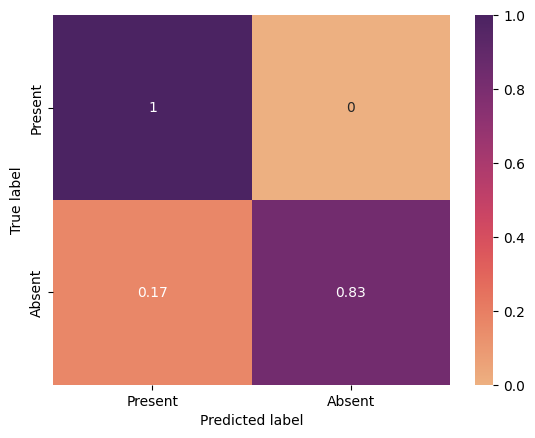}
    \caption{ViT pre-trained for brain MRI data.}
    \label{fig:conf_pre_mri}
\end{minipage}
\begin{minipage}{0.49\textwidth}
   \centering
    \includegraphics[width=0.97\textwidth]{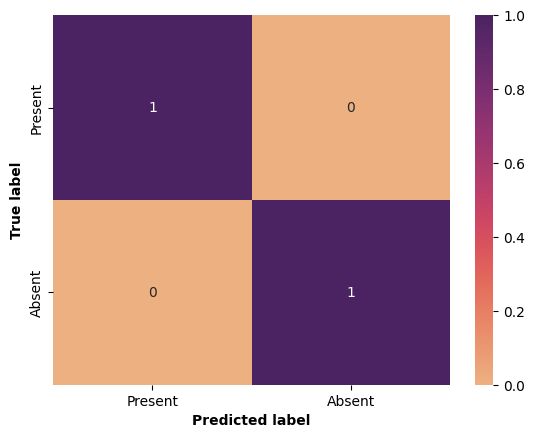}
    \caption{ViT Proposed for brain MRI data.}
    \label{fig:conf_vit_mri_pro}
    \end{minipage}
\end{figure}

The proposed ViT model and the above state-of-the-art models are trained on the training dataset from blood cell images and brain MRI images. All the models are evaluated in terms of classification accuracy, precision, recall, and F-1 score and are tabulated as presented in Table \ref{tab:results_vit}. Classification performances by each of the models are also illustrated using confusion matrices as shown in Fig.'s \ref{fig:conf_cnn}, \ref{fig:conf_alexnet}, \ref{fig:conf_vit_mri_pro}, and \ref{fig:conf_vit_mri_pro}.
\begin{table}[!ht]
  \centering
  \caption{Results summaries for blood image and brain MRI dataset}
  \resizebox{0.65\textheight}{!}{%
    \begin{tabular}{|p{6.035em}|l|c|c|c|c|c|}
    \hline
    Dataset & Metrics & CNN (Custom)  & AlexNet \cite{alexnet} & ResNet18 \cite{resnet50} & ViT \cite{vit}  & \textbf{ViT Proposed} \\
    \hline
    \multicolumn{1}{|l|}{\multirow{4}{*}{Blood Cell Images}} & Accuracy (\%) & 76.20 & 83.31 & 86.80 & 88.38 & \textbf{93.38} \\
    \cline{2-7}          & Precision & 0.80  & 0.86  & 0.90  & 0.90  & \textbf{0.93} \\
    \cline{2-7}          & Recall & 0.77  & 0.84  & 0.87  & 0.88  & \textbf{0.94} \\
    \cline{2-7}          & F-1 Score & 0.77  & 0.83  & 0.87  & 0.89  & \textbf{0.94} \\
    \hline
    \multicolumn{1}{|l|}{\multirow{4}{*}{Brain MRI Images}} & Accuracy (\%) & 64.70 & 88.24 & 92.16 & 92.15 & \textbf{96.15} \\
    \cline{2-7}          & Precision & 0.88  & 0.88  & 0.92  & \textbf{1.00} & 0.99 \\
    \cline{2-7}          & Recall & 0.82  & 0.88  & 0.92  & 0.83  & \textbf{0.99} \\
    \cline{2-7}          & F-1 Score & 0.83  & 0.88  & 0.92  & 0.91  & \textbf{0.97} \\
    \hline
    \end{tabular}}
  \label{tab:results_vit}%
\end{table}%

\begin{figure}[!ht]
   \centering
   \begin{minipage}{0.49\textwidth}
    \includegraphics[width=0.97\textwidth]{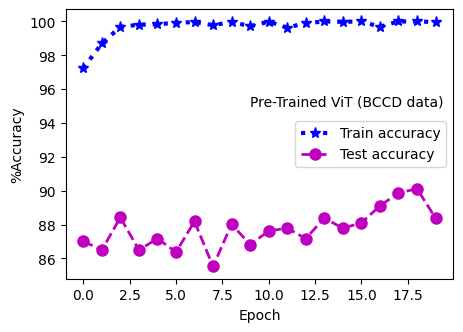}
    \caption{ViT pre-trained for BCCD data.}
    \label{fig:acc_pre}
\end{minipage}
\begin{minipage}{0.49\textwidth}
   \centering
    \includegraphics[width=0.97\textwidth]{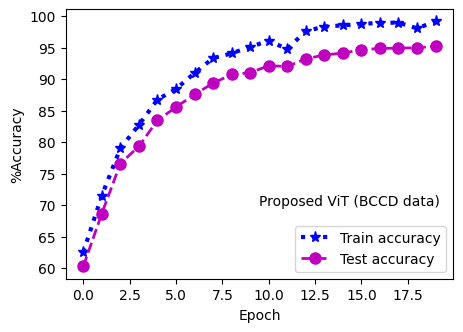}
    \caption{ViT Proposed for BCCD data.}
    \label{fig:acc_pro}
    \end{minipage}
\end{figure}

\begin{figure}[!ht]
   \centering
   \begin{minipage}{0.49\textwidth}
    \includegraphics[width=0.97\textwidth]{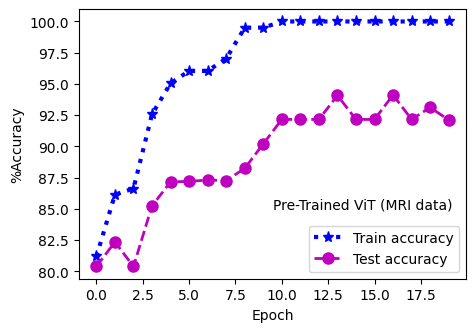}
    \caption{ViT pre-trained for brain MRI data.}
    \label{fig:acc_pre_mri}
\end{minipage}
\begin{minipage}{0.49\textwidth}
   \centering
    \includegraphics[width=0.97\textwidth]{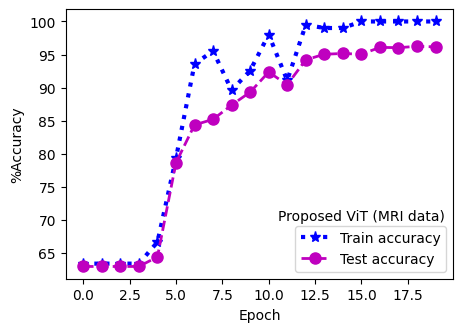}
    \caption{ViT Proposed for brain MRI data.}
    \label{fig:acc_pro_mri}
    \end{minipage}
\end{figure}

\subsection{Discussion} The classification performances of the proposed method and selected state-of-the-art methods lead to the following observations:
\begin{enumerate}
    \item The evaluation metrics comparison in table \ref{tab:results_vit} reveals that the image classification performance is improved by adding the residual of the best attention output selected based on highest L-1 norms of the probabilistic attention score.
    \item Also, the evaluation metrics comparison reveals that the performance by the proposed method is much better than the convolution based models: CNN, AlexNet, and ResNet18.
    \item It can further be observed that all the model's performances are better for the tumor detection using brain MRI data that is for the binary class problem.
    \item The confusion matrices shown in figures \ref{fig:conf_cnn}, \ref{fig:conf_alexnet}, \ref{fig:conf_vit_mri_pro}, and \ref{fig:conf_vit_mri_pro} states the prediction report with respect to the ground truth in terms of fraction of samples. It can be observe red that the classification performance by the proposed framework is better than the pre-trained ViT model for both the dataset cases. To maintain the simplicity of the report, the confusion matrices for other state-of-the-art models are not presented here.
    \item The training and validation accuracies curves are shown in figures \ref{fig:acc_pre}, \ref{fig:acc_pro}, \ref{fig:acc_pre_mri}, and \ref{fig:acc_pro_mri}. It can be observed that the gap between the training accuracy curve and the test accuracy curve are smaller for the proposed method for the both of the datasets considered. Therefore, the proposed framework has more confidence on the validation data.
\end{enumerate}

\subsection{Complexity Analysis} The complexity of the multi-head attention mechanism in traditional vision transformers per layer is $O(n^2 \cdot d)$, where $n$ and $d$ represent the sequence length and dimension of the heads, respectively \cite{vit}. In our proposed framework, we introduce a sorting step for the L-1 norms of attention scores in each layer. Computing the L-1 norms for a matrix of size $n \times n \times d$ has a complexity of $O(n \cdot n \cdot d)$, and sorting the L-1 norms, which has a size equal to the number of heads $d$, has a complexity of $O(d \cdot d)$. Therefore, the total complexity for the ViT with the proposed attention mechanism becomes $O(n^2 \cdot d) + O(n^2 \cdot d) + O(d^2)$. By ignoring the summation term with lower complexity, the overall time complexity of the proposed framework remains almost the same as that of the traditional ViT.

\section{Conclusions}\label{ch_vit:conclusion}  In this chapter, we propose a novel framework of multi-head self-attention for the Vision Transformer (ViT) encoder block. The proposed framework utilizes the L-1 norms of attention scores to identify the best attention output. The attention scores are the SoftMax of the scaled dot product of Query and Key.  In the proposed mechanism, the attention scores are sorted based on L-1 norms to obtain the highest attention score and the corresponding attention head index. The output of that particular head is added to the final projected attention at the end of the linear layer. The aggregated attention output is then given to the further parts of the encoder block (layer norms and feed-forward layers).

To evaluate the performance of the proposed ViT framework, we conducted experiments on two datasets: (i) blood cell images for blood cell type classification, and (ii) Brain MRI images for brain tumor detection. The experimental results demonstrate that the proposed framework achieves superior classification performance compared to both the traditional ViT model and state-of-the-art convolution-based methods on both datasets.

In the current research work, we integrate the proposed attention mechanism with a pre-trained ViT base model, replacing the existing attention block. Subsequently, the resulting model is fine-tuned with the datasets considered. Future research can extend this work by training the ViT model from scratch using the proposed attention framework on large datasets.

\medskip

\bibliographystyle{IEEEtran.bst}
\bibliography{Reference.bib}

\end{document}